# Lessons Learnt from Field Trials of a Robotic Sweet Pepper Harvester for Protected Cropping Systems

**Christopher Lehnert**[*1], Christopher McCool[1], Tristan Perez[1]
[1] School of Electrical Engineering and Computer Science, Queensland University of Technology, Brisbane, Australia.

**Abstract**
In this paper, we present the lessons learnt during the development of a new robotic harvester (Harvey) that can autonomously harvest sweet pepper (capsicum) in protected cropping environments. Robotic harvesting offers an attractive potential solution to reducing labour costs while enabling more regular and selective harvesting, optimising crop quality, scheduling and therefore profit. Our approach combines effective vision algorithms with a novel end-effector design to enable successful harvesting of sweet peppers. We demonstrate a simple and effective vision-based algorithm for crop detection, a grasp selection method, and a novel end-effector design for harvesting. To reduce complexity of motion planning and to minimise occlusions we focus on picking sweet peppers in a protected cropping environment where plants are grown on planar trellis structures. Initial field trials in protected cropping environments, with two cultivars, demonstrate the efficacy of this approach. The results show that the robot harvester can successfully detect, grasp, and detach crop from the plant within a real protected cropping system. The novel contributions of this work have resulted in significant and encouraging improvements in sweet pepper picking success rates compared with the state-of-the-art. Future work will look at detecting sweet pepper peduncles and improving the total harvesting cycle time for each sweet pepper. The methods presented in this paper provide steps towards the goal of fully autonomous and reliable crop picking systems that will revolutionise the horticulture industry by reducing labour costs, maximising the quality of produce, and ultimately improving the sustainability of farming enterprises.

**Background**
Autonomous harvesting has been demonstrated on a number of horticultural crops such as sweet peppers (Hemming et al. 2014; Bontsema et al. 2014), cucumbers (Henten & Hemming 2002), citrus fruits (Mehta & Burks 2014), strawberries (Hayashi et al. 2010) and apples (Bulanon & Kataoka 2010; De-An et al. 2011).

Autonomous sweet pepper harvesting has previously been demonstrated by the Clever Robots for Crops (CROPS) project (Hemming et al. 2014; Bontsema et al. 2014). This work shows that there is still further research and development required to solve the challenges associated with robotic harvesting of sweet peppers.

Detection and segmentation are essential in knowing the approximate location of the crop within an image and approaches have been developed for both automated harvesting and yield estimation (Mccool et al. 2016; Perez et al. 2015). A common technique for robotic crop harvesting is to perform crop detection and segmentation using 3D information (position, orientation and shape) using sensors such as a structured-light camera (Lehnert et al. 2016) or stereo vision (Henten & Hemming 2002). This 3D localisation step also can be used to determine a grasp and/or cutting pose. Grasp locations can be determined by using prior geometric models fitted to the 3D scene or by using a heuristic to rank local surface information of the segmented 3D crop (Lehnert et al. 2016; Lehnert et al. 2017; Sa et al. 2016).

A range of manipulator configurations have been used for autonomous harvesting projects including and several works have compared various joint configurations to optimise target-reachability in cluttered environments (Van Henten et al. 2009; Lehnert et al. 2015).

One key component for a robotic harvester is the end-effector that grasps and/or cuts the crop. A common solution is to use suction cups (Hayashi et al. 2010; Van Henten et al. 2003) usually because of their mechanical simplicity and minimal surface area required to hold onto the crop. Another gripping alternative are contact-based grippers (De-An et al. 2011; Ling et al. 2005) which generally employ mechanical fingers that close around the crop. Contact-based grippers can grip the crop very securely, but are more prone to interference from nearby objects such as branches and other crops. Some crops such as sweet pepper and cucumber must also be cut from the plant and so require an additional detachment tool such as a thermal cutter (Henten & Hemming 2002) or scissor-like mechanism (Hemming et al. 2014; Hayashi et al. 2010; Han et al. 2012). End effectors customised to a particular crop are common, for example in Hemming et al. 2014 developed a custom harvesting tool which simultaneously envelops and cuts-free sweet peppers with a hinged jaw mechanism. This mechanism was found to be more effective than a scissor mechanism, however size and geometry constraints restricted access to some sweet peppers (Hemming et al. 2013).

**Initial Prototypes**
Among the many solutions for harvesting end effectors this work has investigated alternative strategies that could be utilised for robotic crop harvesting. These prototypes were investigated in both an outdoor field and indoor protected cropping environment. The lessons learnt from these prototypes are presented in this section

Field-grown sweet peppers grow in rows of short shrubs, where the crop is mostly located within the 3D structure of the plant. This makes identifying sweet peppers using vision challenging. Successful picking would also likely require sensing the structure of the plant and performing path planning and manipulation in a dense and deformable environment. An example of an outdoor environment is shown in Figure 1 (b) and (c). Direct sunlight and deep shadows also caused issues for both depth cameras and colour cameras to become saturated in parts of the image. To accelerate progress and improve the chances of success, development was subsequently focused on working in protected cropping systems, where many of these challenges are reduced or eliminated.

Before deciding on the harvesting tool tested in the field trials described in the following sections (see Figure 2 (b)), we experimented with many alternative gripping and detachment methods, illustrated in Figure 1. Initial testing included using two robots, Baxter and Kinova Mico (see Figure 1 (a) and (b)) that were available immediately in our lab. Both robots were used to test assumptions and demonstrate initial proof-of-concept perception and motion planning algorithms.

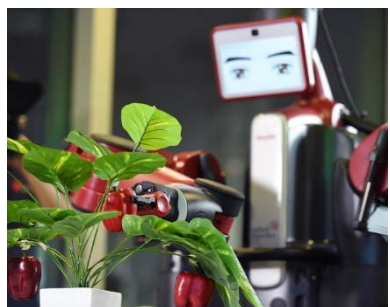
(a) Baxter prototype

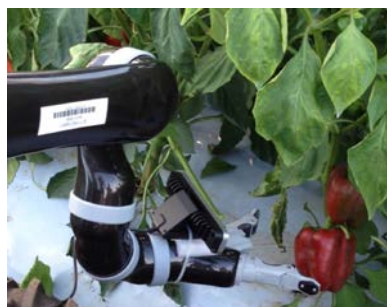
(a) Kinova Mico prototype

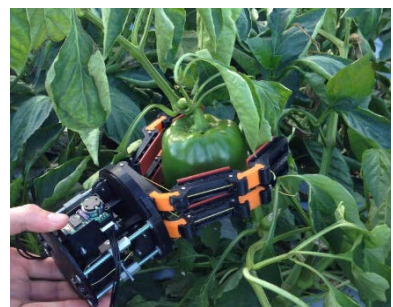
(c) 3D printed gripper

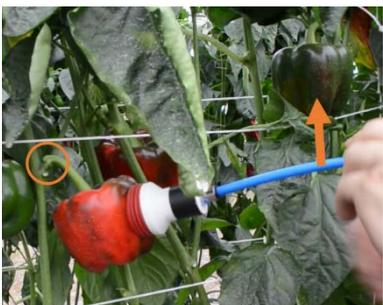
(d) Snapping prototype using suction cup

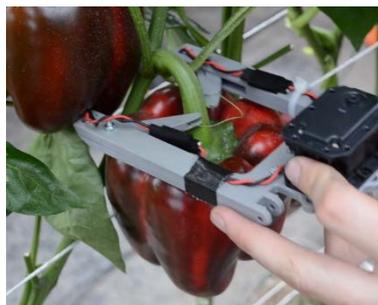
(e) wire cutter prototype – before attachment

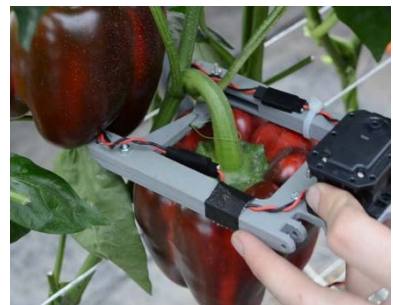
(f) wire cutter prototype – after attachment

*Figure 1 - (a) Initial software development using Baxter and Kinect v2 sensor in a lab environment. (b) Initial field testing using lightweight Kinova Mico robot arm and eye-in-hand RGB-D camera. (c) 3D printed under-actuated finger gripper tested on field-grown sweet peppers. (d). Detaching sweet peppers using suction cup and flexible hose. Pulling gently upwards on the flexible hose causes a clean break at the abscission zone (circled). (e) Prototype wire cutting mechanism that is pushed onto the peduncle through a set of spring-loaded fingers to wrap a flexible cable around the peduncle. A servo motor then pulls the cable tight to cut the peduncle.*

A 3D printed under-actuated four finger gripper shown in Figure 1 (c) was constructed and tested. This gripper was adapted from the open source gripper design (Dollar & Howe 2010). The design of the gripper was modified to suit the nominal size and shape of sweet peppers using an optimisation approach determining the best finger link lengths to grasp different sizes of sweet peppers. This gripper was tested in the field (see Figure 1 (c)) on real sweet peppers by hand. The main disadvantage we found with this gripper was while grasping around the sweet pepper, neighbouring branches or leaves would be caught in the grasp. Furthermore, as the fingers were positioned in parallel the sweet pepper tended to pivot when forces were applied to the crop causing it to slip out of the fingers.

Removing sweet peppers without a cutting mechanism by pulling/snapping them from the plant was also investigated. Human workers within field-grown sweet peppers generally pick with a motion that rotates the sweet pepper around its peduncle causing a clean break at the abscission zone (a layer of cells near the end of the peduncle where the crop naturally separates when ripe). Detaching sweet peppers at the abscission zone fetches a higher market price, leads to faster plant recovery and less susceptibility to diseases. A detachment method that we found to be relatively successful and simple to implement was to pull upwards on the sweet pepper using a suction cup via a flexible tether. Initial testing of this method was performed by pulling upwards on a suction cup attached to a flexible vacuum hose (as shown in Figure 1 (d)). The strategy was able to remove 17 out of 22 sweet peppers (14 cleanly and 3 with some tearing of the peduncle). This strategy was found to be most reliable on longer thinner peduncles whereas thicker peduncles on overly mature sweet peppers tended to tear rather than detach cleanly.

Another mechanism investigated was the "wire cutter" shown in Figure 1(e) and (f), which cuts peduncles by tightening a loop of thin flexible wire around them. The mechanism is first moved forward so that the peduncle pushes past a spring-loaded set of one-way fingers that close behind it. A loop of wire that protrudes from a small hole in the tip of each finger now encapsulates the peduncle. A servo motor then tightens the wire until it slices through the peduncle. Micro switches give feedback on when the fingers have opened and closed behind the peduncle. This type of cutting mechanism is relatively robust to errors in the estimated peduncle position, and cannot inadvertently cut the crop, since the wire can only cut objects small enough to pass through the finger mechanism. A disadvantage of this type of cutting mechanism is the need to have mechanical parts protrude past the peduncle to latch onto it, which means branches and crop nearby may interfere with the operation of the cutter. This mechanism was tested in a protected cropping system and failed to detach sweet peppers with tough/thick peduncles and therefore determined unfeasible in the final design.

**Harvesting Method**
Our method for automated harvesting consists of 5 main stages including: scanning, detection, grasp selection, attachment and detachment. More details of each stage of the harvesting method can be found in the following prior work (Lehnert et al. 2016; Mccool et al. 2016; Lehnert et al. 2017; Sa et al. 2016). Figure 2 (a) shows Harvey the robotic sweet pepper harvester deployed in a real protected cropping system and features a custom mobile base, 7 Degree of Freedom robot arm and a custom harvesting end effector.

Scanning is the process of moving the vision sensor in the correct path using the robotic manipulator to build up enough information about the scene. The scanning path can involve various options such as a path that moves around an individual crop of interest or a path that encompasses multiple crops. To save harvesting time per crop this stage can be simplified to moving to a single location which captures a view of multiples crops. Detection is the stage which determines the location of the crop from the information built up from the scanning process. For sweet peppers this is achieved by using colour information. The next stage of the harvesting process is to determine the grasp and cutting locations relative to the detected crop. Two methods have been tested, a model based method (Lehnert et al. 2016) and a heuristic based methods using 3D surface information (Lehnert et al. 2017). Once a grasp and cutting location is determined the final stage is for the robot to attach the suction cup to the target crop and remove the crop from the plant using a cutting tool. For this work the suction cup and cutting tool have been designed to be separated using a magnet in order to place the suction cup and cutting tool at two different locations on the crop and the design is shown in Figure 2 (b).

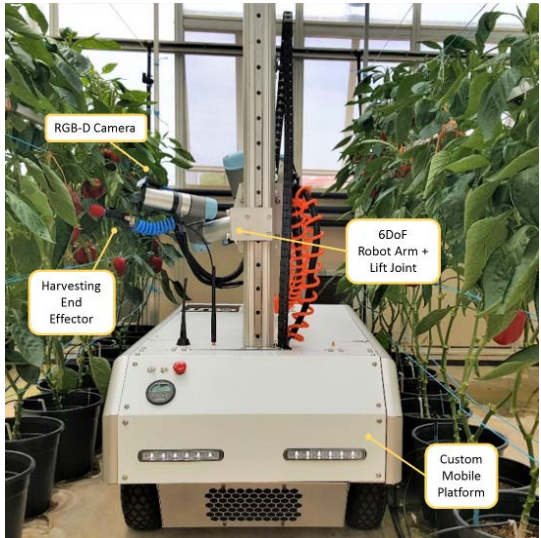
(a) Custom mobile base platform

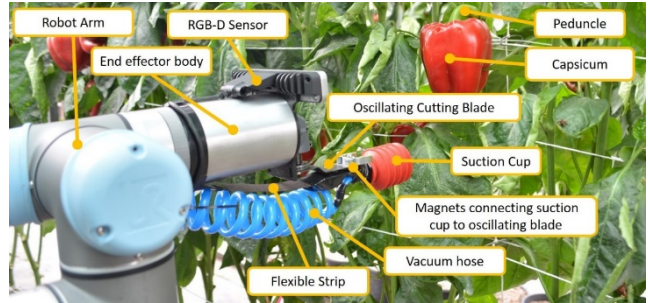
(a) Custom harvesting tool

*Figure 2 - Harvey the robotic sweet pepper harvester and labelled system components. (a) Image of the harvesting platform in a protected cropping system highlighting the base, robot arm and harvesting end effector. (b) A view of the custom harvesting tool on the end of the robot arm featuring a suction cup, cutting tool, 3D camera and magnetic separation mechanism.*

**Results**

Two field trials were conducted over two harvesting seasons on a farm in North Queensland (Australia) in a protected cropping system. The first trial involved an initial prototype of the robot on a scissor-lift base whereas the second field trial involved the custom mobile base platform seen in Figure 2 (a). The results from both field trials are given in Table 1. These results compare the attachment, detachment rates of the robotic system for two scenarios, modified or unmodified. For the modified scenario, the common practice of leaf removal was conducted before the trial in order to improve the visibility of the crop. The unmodified scenario evaluated the robotic harvester without leaf removal. Out of the total sweet peppers, 14/24 (58%) and 11/26 (42%) were successfully harvested for field trial 1 and 2 respectively, where a successful harvest includes a successful attachment and detachment. During field trial 1 and 2, a total of 22/24 (92%) and 11/26 (42%) sweet peppers were successfully detached from the plant, irrespective of an attachment success. On the other hand, attachment rates of 14/24 (58%) and 21/26 (81%) were achieved for field trial 1 and 2, respectively.

*Table 1 - Field Trial Harvesting Results*

|  | Trial 1 – Cultivar Claire | Trial 2 – Cultivar Redject |
|---|---|---|
| Modified Detachment Rate | 92% | 42% |
| Modified Attachment Rate | 58% | 81% |
| Modified Attachment & Detachment | 58% | 42% |
| Unmodified Attachment & Detachment | 46% | N/A |
| Avg. Harvesting time | 34-40 sec | N/A |

**Discussion & Conclusion**

This paper describes an autonomous crop harvesting system that achieves state-of-the-art results for sweet pepper harvesting in a protected cropping environment. We highlight some initial prototypes which offer key insights into some of the challenges involved in automated harvesting using a robotic system.

The results demonstrate that visual detection of the crop can be achieved in very difficult situations. The system is demonstrated using a custom end-effector that uses both a suction gripper and oscillating blade to successfully remove sweet peppers in a protected cropping environment. The presented harvesting system achieved a 58% success rate taking into account modifications such as leaf removal, which improves upon the previous state-of-the-art systems by almost twice the performance. In addition, we demonstrate a success rate for the unmodified scenario of 46%. Despite these significant improvements over the state of the art, a number of issues were made apparent throughout our experiments.

The perception and planning systems have been shown to perform well for our specific problem. However, further advancements are necessary for a general system. To achieve a generic visual crop detection system suitable for a range of crops, we believe a fast learning system capable of high detection accuracy that can be trained from a small number of training images is required. This is a challenging task which remains unsolved. In future work, we will investigate the quality of the visual features presented in this paper under more challenging conditions.

On the hardware side, the most common detachment failure case was found to be the cutting tool missing either side of the peduncle. In this work, we have no explicit detection and localisation of the peduncle. To improve the detachment reliability, future work will be aimed at detecting and localising the peduncle in which preliminary results can be found in (Sa et al. 2016).

The harvesting cycle time is currently slow (approximately 33 seconds), and reducing this cycle time is part of future work. One of the most time-consuming processes is the scanning step. Possible steps to improve this is to scan over a larger area incorporating multiple sweet peppers or only detect crop from a single view. The movement of the robot arm will also be sped up significantly in future work to further reduce cycle times.

The novel contributions of this work have resulted in significant and encouraging improvements in sweet pepper picking success rates compared with the state-of-the-art. We believe that continuing to build on this initial system presented in this paper will result in further meaningful progress towards a commercially viable system. The methods presented in this paper may also be applied to a range of other high-value horticultural crops, and provides steps towards the goal of fully autonomous and reliable crop picking systems to revolutionise the horticulture industry by reducing labour costs, maximising the quality of produce, and ultimately improving the sustainability of farming enterprises.

**Acknowledgments**
This project has received funding from the Queensland Department of Agriculture and Fisheries and QUT Strategic Investment in Farm Robotics Program. The Authors would also like to acknowledge contributions from Andrew English, Inkyu Sa and Raymond Russel while they were employed within the Strategic Investment in Farm Robotics program at QUT.